\documentclass[10pt,twocolumn,letterpaper]{article}

\usepackage[pagenumbers]{cvpr} 

\usepackage[dvipsnames]{xcolor}

\usepackage{graphicx}
\usepackage{amsmath}
\usepackage{amssymb}
\usepackage{booktabs}

\usepackage{algorithm}
\usepackage{algorithmic}
\usepackage{amsmath,amssymb}
\usepackage{subcaption}

\usepackage{mathtools}
\usepackage{booktabs}
\usepackage{multirow}
\usepackage{float}
\usepackage{graphicx,wrapfig,lipsum}
\usepackage[symbol]{footmisc}

\definecolor{cvprblue}{rgb}{0.21,0.49,0.74}
\usepackage[pagebackref,breaklinks,colorlinks,citecolor=cvprblue]{hyperref}

\def\paperID{10342} 
\def\confName{CVPR}
\def\confYear{2024}

\title{FedUV: Uniformity and Variance for Heterogeneous Federated Learning \footnotemark[1]}

\author{Ha Min Son\\
University of California, Davis\\
\and
Moon-Hyun Kim\\
Hippo T\&C\\
\and
Tai-Myoung Chung\\
Hippo T\&C\\
\and
Chao Huang\\
University of California, Davis\\
\and
Xin Liu\footnotemark[2]\\
University of California, Davis\\
}

\begin{document}
\maketitle
\footnotetext[1]{To be presented at CVPR 2024}
\footnotetext[2]{Correspondence to xinliu@ucdavis.edu}
\begin{abstract}
Federated learning is a promising framework to train neural networks with widely distributed data. However, performance degrades heavily with heterogeneously distributed data. 
Recent work has shown this is due to the final layer of the network being most prone to local bias, some finding success freezing the final layer as an orthogonal classifier. 
We investigate the training dynamics of the classifier by applying SVD to the weights motivated by the observation that freezing weights results in constant singular values. We find that there are differences when training in IID and non-IID settings.
Based on this finding, we introduce two regularization terms for local training to continuously emulate IID settings: (1) variance in the dimension-wise probability distribution of the classifier and (2) hyperspherical uniformity of representations of the encoder. These regularizations promote local models to act as if it were in an IID setting regardless of the local data distribution, thus offsetting proneness to bias while being flexible to the data.
On extensive experiments in both label-shift and feature-shift settings, we verify that our method achieves highest performance by a large margin especially in highly non-IID cases in addition to being scalable to larger models and datasets.
\end{abstract}    
\section{Introduction}
\label{sec:intro}

Federated Learning (FL) \cite{fedavg_mcmahan2017} is a distributed learning framework that allows the training of deep neural networks with data in decentralized locations. FL is especially appealing because it achieves similar performance to centralized training in specific settings while also negating the cost of collecting data into a centralized location and allowing effective parallelization across devices \cite{verbraeken_distributed}. However, because devices participating in FL use locally collected data for training, it is realistic to expect that the collective data across all devices are not independent and identically distributed (non-IID). This severely degrades overall performance. Local devices optimize their copy of the model towards local optima. In non-IID settings, however, it is likely that local optima across devices are in disagreement with the true optimum of the IID setting. Thus, the direction of the gradient causes dissonance and degraded performance. This phenomena is referred to as \textit{client drift} in the FL literature.

To combat client drift, a number of work \cite{fedprox_li2020federated, feddyn_acar2021federated, moon_li2021model} introduced the idea of using the global model as a basis for regularization. More recently, some work showed it may be effective to focus on certain layers which are more prone to bias. In particular, \cite{ccvr_luo2021no} found that the final layer, referred to as the \textit{classifier}, is the most biased in non-IID settings. They introduce a method to calibrate the classifier to offset this bias. Other work \cite{fedalign_mendieta2022local} used augmentation techniques \cite{yang2020gradaug} to regularize the classifier, while another work \cite{fedbabu_oh2021fedbabu} randomly initialized and froze the weights of the classifier. 
Since random vectors in high dimension space are likely to be orthogonal, the output of the penultimate layer, referred to as the \textit{representations} or \textit{activations} of the encoder, are trained to fit these orthogonal classifiers, thus offsetting bias.

However, we note previous work have approached the non-IID problem by regularizing local models to be \textit{less biased}, rather than directly regularizing local models to \textit{emulate the IID setting}. In addition, many work have not been tested in a feature-shift setting, often only using the Dirichlet distribution to simulate a label-shift setting.
Furthermore, while providing important insight for FL, these work generally do not take into account efficiency and scalability. State-of-the-art methods such as FedProx \cite{fedprox_li2020federated}, SCAFFOLD \cite{scaffold_karimireddy2020scaffold}, and FedDyn \cite{feddyn_acar2021federated} require distances calculations of weights between all layers of multiple models at every batch. 
MOON \cite{moon_li2021model} and FedAlign \cite{fedalign_mendieta2022local} require extra forward passes at every batch. 
Computation and memory constraints certainly become issues as the size of models and datasets grow.

\begin{figure}
    \includegraphics[width=0.95\columnwidth,trim=4 1 4 4,clip]{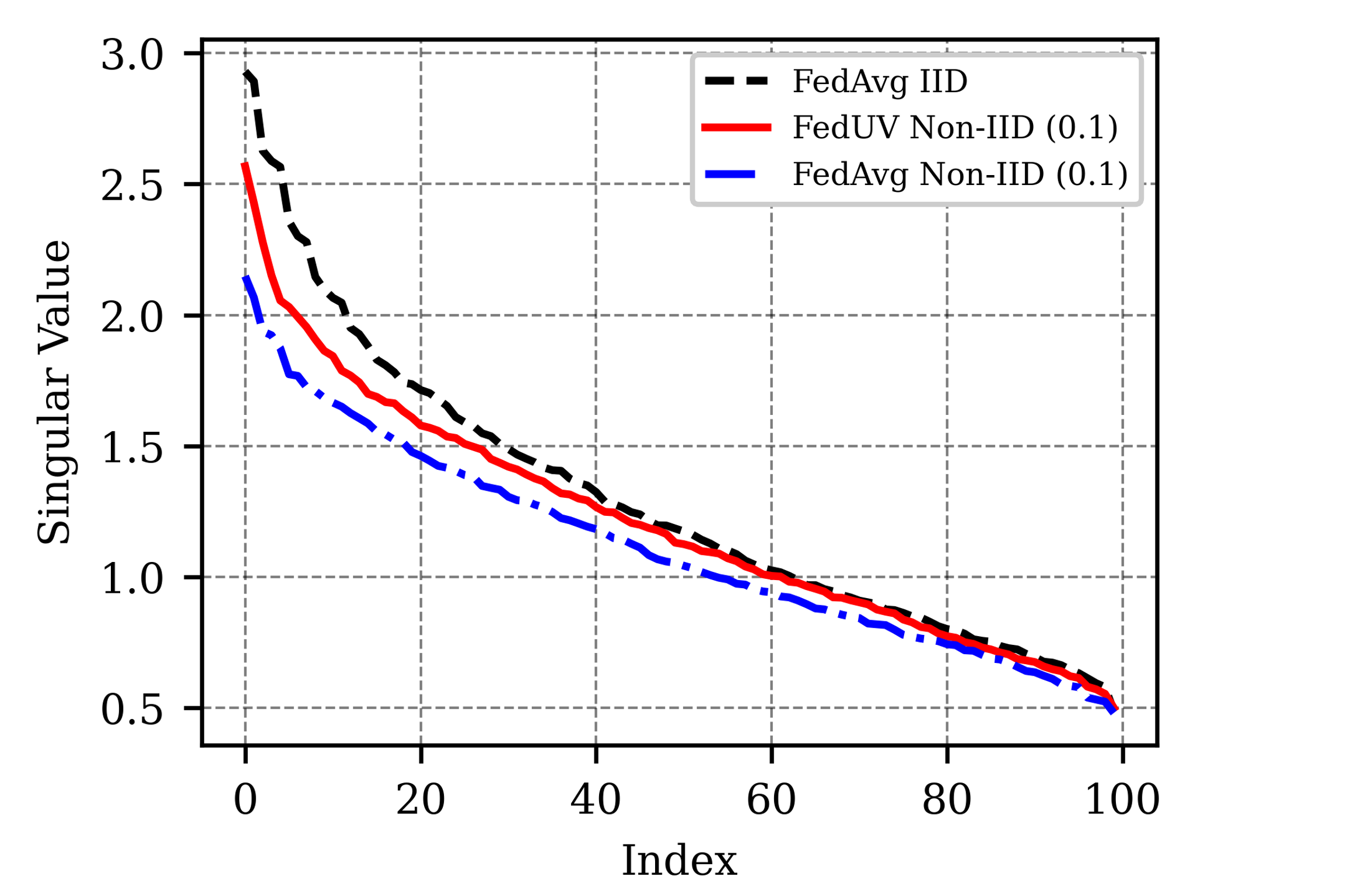} 
\caption{Singular values of the weights of the classifier (final layer) trained on CIFAR-100. Training setup is specified in Section \ref{experimental_setup}}
\label{fig_svdcomp}
\end{figure}

\noindent \textbf{Motivation.}
In most classification problems, it is standard to minimize a Cross-Entropy loss function \cite{CELoss_good1992rational} to fit a neural network's probability distribution to the sampled probability distribution. 
When the sampled class distribution is imbalanced, such as in a non-IID environment, the classifier becomes biased towards some subspace regardless of the features extracted by the encoder.  
Motivated by the success of freezing the classifier as orthogonal vectors \cite{fedbabu_oh2021fedbabu}, we use the singular value decomposition to gain insight into the training dynamics of the classifier.
We train a Resnet-18 \cite{resnet_he2016deep} on CIFAR-100 \cite{dataset_cifar10_100} and take the singular value decomposition of the weight matrix ($\mathbf{W} = \mathbf{U}\mathbf{\Sigma}\mathbf{V^\top}$, $\mathbf{\Sigma} = diag(\mathbf{\sigma_{1},...,\sigma_{n}})$) as shown in Fig.~\ref{fig_svdcomp}.
The singular values of the classifier decrease in a non-IID setting, showing there is a difference in the training dynamic between IID and non-IID settings.

In light of this, we propose an approach to the non-IID problem in which we directly promote the emulation of an IID setting. There is a nuanced difference from previous work which regularize to decrease bias. Previous work use the global model as a source of regularization under the intuition that the global model is less biased than any single local model. However, this may not be optimal in every setting as the aggregate of biased local model may still be biased. We instead focus on regularizing local models to emulate the IID setting without the use of the global model.
Namely, we present FedUV, a method that promotes IID emulation by inducing representation hyperspherical \textbf{U}niformity and classifier \textbf{V}araiance during local training. FedUV penalizes classifier probability distributions that are biased towards a subset of classes, thus promoting probability distributions in non-IID settings to match IID settings. This promotes local classifiers to be unbiased in their predictions rather than becoming biased towards local labels. Furthermore, we penalize representations that are not thoroughly spread across the hypersphere. There are two reasons for this. First, this encourages the encoder to not become biased towards local features. Second, expanding the feature space allows the classifier to expand its variance in more directions.  We show both regularizations are essential for improved performance. FedUV achieves state-of-the-art performance on various standard label-shift benchmark in addition to various feature-shift benchmarks while being extremely simple, efficient, and scalable.

Our contributions are summarized as follows. First, we verify that singular values of the classifier decreases as a consequence of non-IID environments in FL (Fig.~\ref{fig_svdcomp}). 
Second, we present two regularization terms, representation hyperspherical uniformity and classifier variance, to prevent this degradation. These regularization terms are simple yet effective.
Third, we show that FedUV achieves state-of-the-art performance not only on label shift non-IID settings but also on feature shift non-IID settings. 
Unlike previous work, FedUV does not use the global model as a source of regularization, which requires extra forward-pass, requires extra memory, or requires matrix comparison between weights of each layer, thus being more efficient.
\section{Related Work}
\label{sec:related_works}

\subsection{Federated Learning (FL)} The basic FL algorithm is FedAvg \cite{fedavg_mcmahan2017}. This algorithm progresses with the repetition of four steps. 
First, (step 1) a server broadcasts its global model. 
Second, (step 2) clients receive and train the model using their own local data.
Third, (step 3) clients upload their trained local model.
Fourth, (step 4) the server aggregates local models to create the next generational global model. 


There are a wide range of recent work in the field of FL such as client selection \cite{clselect_0, clselect_1}, data sharing \cite{sharing_0}, privacy \cite{privacy_0, privacy_1}, communication efficiency \cite{efficient_0, efficient_1, efficient_2}, medical applications \cite{medical_0}, and knowledge distillation \cite{distill_0, distill_1}. As we cannot hope to cover all work on FL, we mainly focus on work that address the performance degradation in non-IID settings. Work in this domain can generally be divided into two groups: 1) papers that focus on the aggregation process (step 4), and 2) efforts that focus on the local training process (step 2). 

\subsection{Aggregation Regularization} Work that focus on aggregation generally do not modify the local training scheme of FedAvg. Clients train their local model on local data using a SGD based optimizer. To address the non-IID problem, they change how the weights of the local model are aggregated. 
FedAvgM \cite{fedavgm_hsu2019measuring} applied a momentum term when aggregating and FedNova \cite{fednova_wang2020tackling} normalized local models before averaging. 
Other work in Personalized FL \cite{pfedLA_ma2022layer, shamsian2021personalized} used hypernetworks to generate the parameter of other client layers. PFNM \cite{fedmaORIG_yurochkin2019bayesian} and FedMA \cite{fedma_wang2020federated} introduced a Bayesian method to align potentially mixed neurons due to the permutation invariance of neural networks. PAN \cite{efficientFedMA_li2022federated} improved efficiency by encoding position into neurons. FedDF \cite{feddf_lin2020ensemble} utilized generated and unlabelled data. CCVR \cite{ccvr_luo2021no} noted the largest bias exists in the classifier (the last layer) than any other layer. They used a Gaussian mixture model to create artificial data and calibrated the classifier.

\subsection{Local Training Regularization} Work that focus on local training generally do not modify the aggregation scheme of FedAvg. 
To address the non-IID problem, they instead penalize local clients based on different criteria. As our work also belongs to this group, we discuss them in more detail. 

FedProx \cite{fedprox_li2020federated} added a ${L_{2}}$ regularizer between local model weights and global model weights that prevent local model weights from becoming dissimilar to the global model. 
FedDyn \cite{feddyn_acar2021federated} added an inner product regularization term between the current round local model and previous round local model noting that regularizing solely based on the global model may cause convergence problems for local models. MOON \cite{moon_li2021model} added a contrastive loss regularization term noting there should be a balance between the representational similarity between global model, the current round local model, and previous round local model.

These methods use the global model as a source of regularization. The intuition is that the global model is less biased than any single local model. However, there are two large issues concerning this approach. First, the global model can also be biased if the local models are biased, since it is an averaged model. This bias may slow convergence and not improve performance. Second, efficiency and scalability becomes an issue. FedProx and FedDyn require the calculation of the ${L_{2}}$ distance between global model and local model, and MOON requires three forward passes at every batch.

More recently, FedAlign \cite{fedalign_mendieta2022local} noted that it is possible to use only the local model for effective regularization. Local models employ GradAug \cite{yang2020gradaug} to regularize itself rather than relying on the global model. However, FedAlign still suffers in efficiency due to its reliance on an additional forward pass and estimation of second order information. FedBABU \cite{fedbabu_oh2021fedbabu}, on the other hand, randomly initializes and freezes the classifier to prevent bias. While this work can be categorized as Personalized Federated Learning, we include FedBABU in our discussion due to its focus on the classifier. We further explore these ideas in Section \ref{actual_results}.

\begin{figure}[t]
\centering

\begin{subfigure}{0.95\columnwidth}{
    \includegraphics[width=\columnwidth,trim=2 7 5 7,clip]{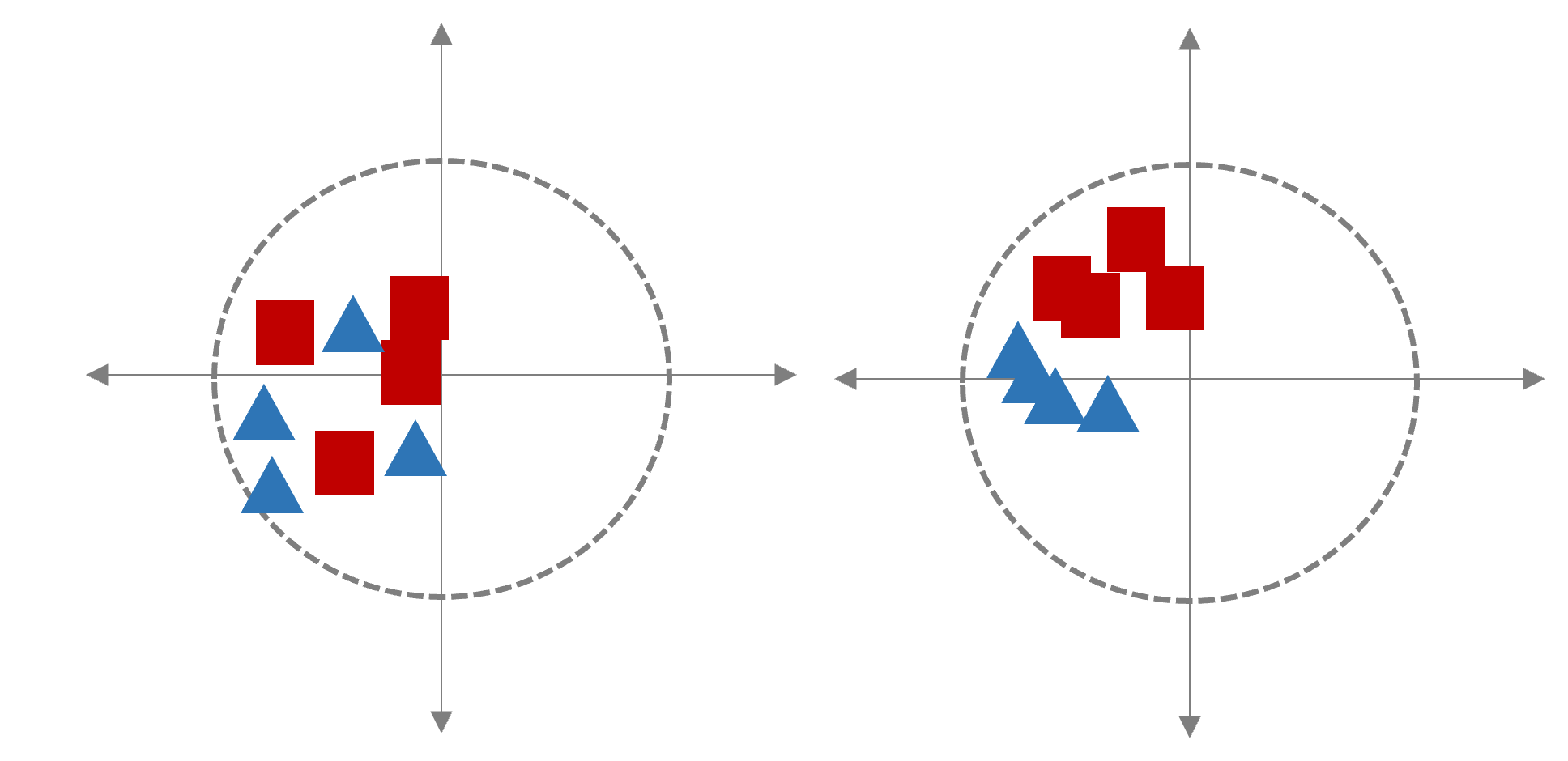} 
    \caption{Cross-entropy loss aligns the representations}
}
\end{subfigure}
\par\bigskip
\begin{subfigure}{0.95\columnwidth}{
    \includegraphics[width=\columnwidth,trim=2 7 5 7,clip]{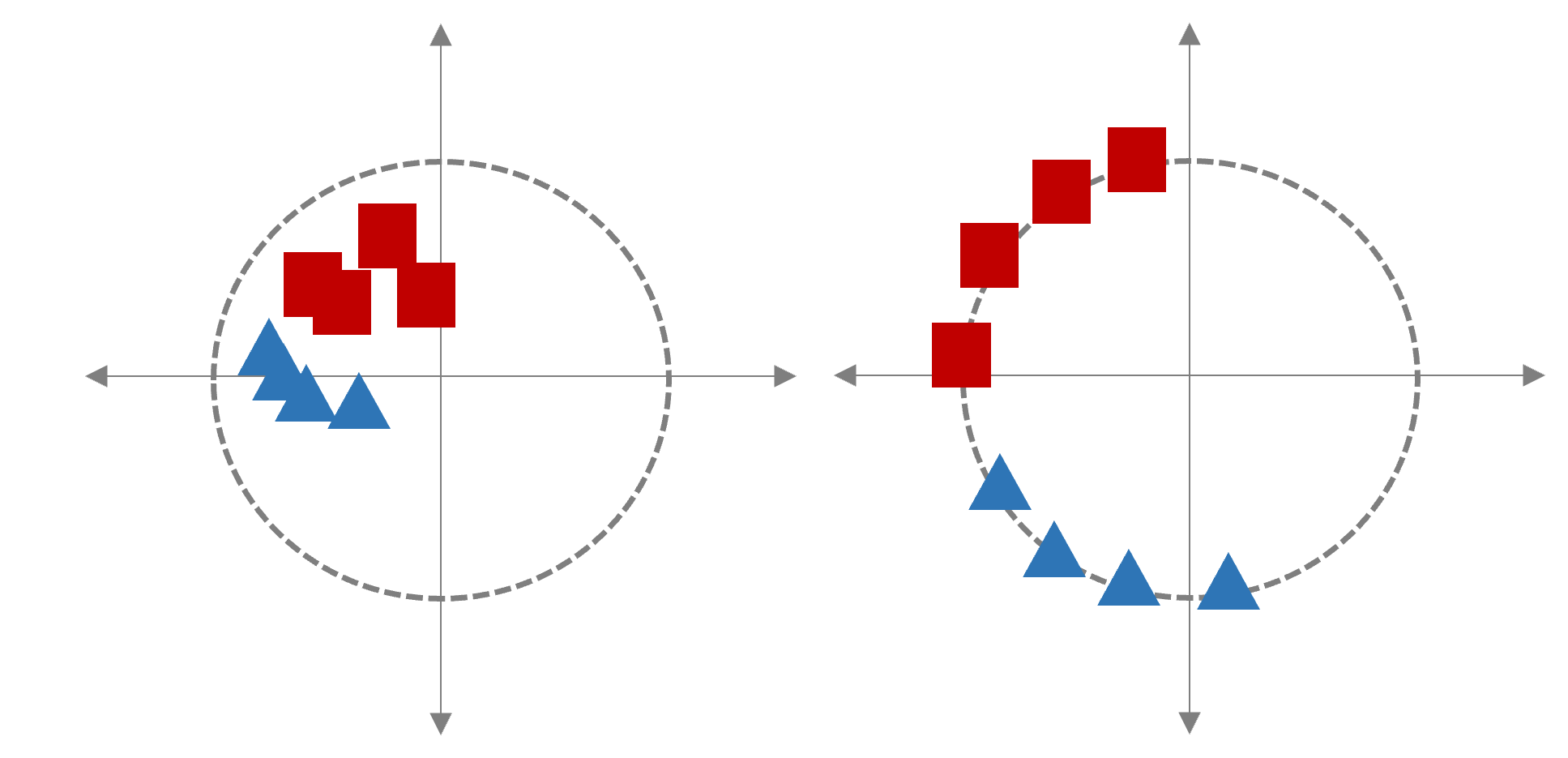} 
    \caption{Hyperspherical Uniformity regularizes the representations (penultimate layer output)}
}
\end{subfigure}
\par\bigskip
\begin{subfigure}{0.95\columnwidth}{
    \includegraphics[width=\columnwidth,trim=2 7 5 7,clip]{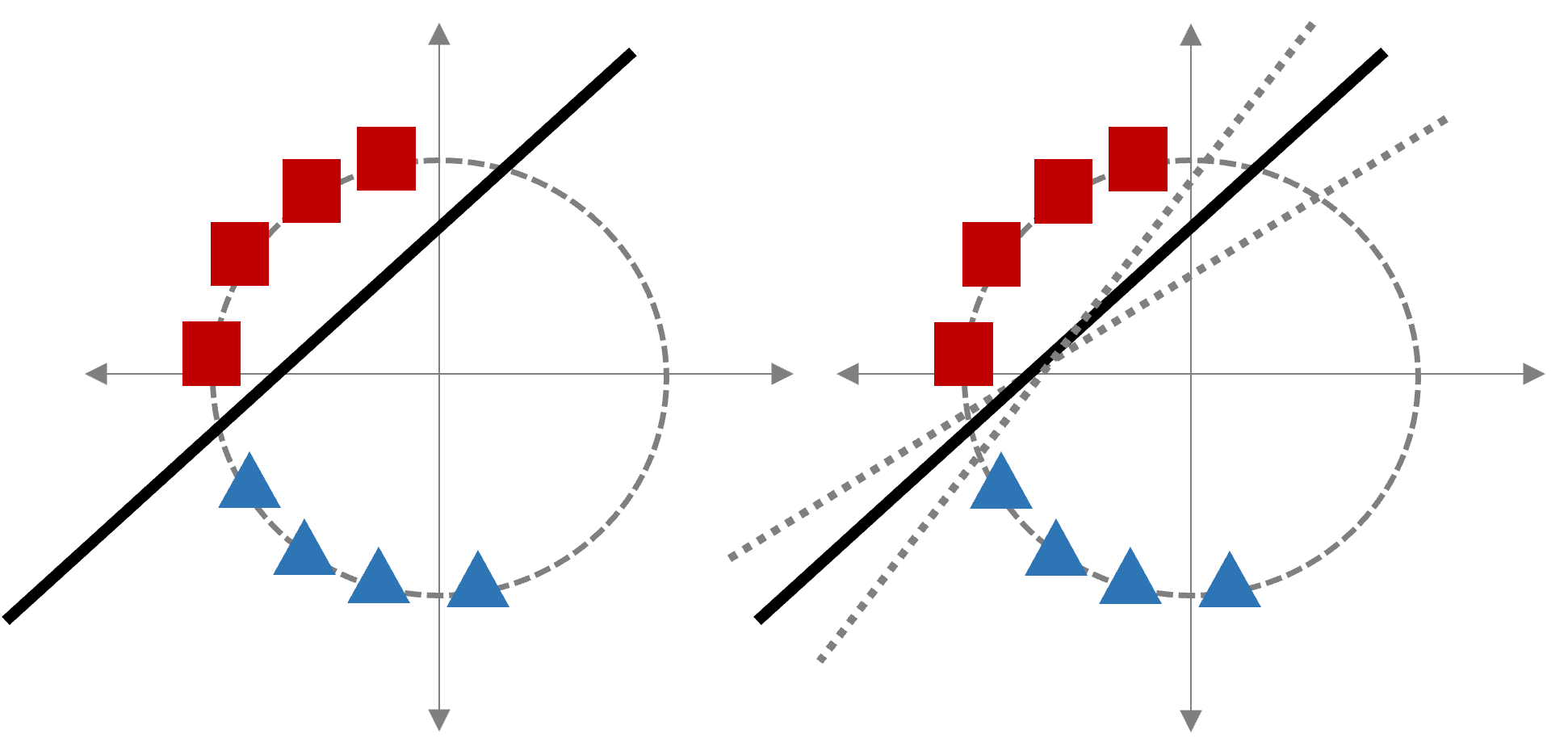} 
    \caption{Variance regularizes the classifier (final layer output)}
}
\end{subfigure}

\caption{Uniformity regularization is applied to the penultimate layer and variance regularization is applied to the output layer.
}
\label{fig_feduv_method}
\end{figure}
\section{Proposed Method --- FedUV}
\label{sec:methods}

The motivation of FedUV comes from two main observations. First, the similarity between local models decreases further in later layers of the model \cite{ccvr_luo2021no}, with the classifier (the final layer) being most biased towards local data. Second, as shown in Fig.~\ref{fig_svdcomp}, the singular values of the classifier of the global model degenerate in non-IID settings. 
Our goal then is to prevent this degeneration, promoting local models to emulate the IID setting regardless of local data distribution. 

We achieve this by penalizing the classifier if it does not act as it would in an IID setting. We use a hinge loss to regularize the classifier if the class-wise variance of the probability distribution does not exceed a constant threshold $c$, which is set to be the class-wise variance of the ground truth labels in a batch with balanced classes. However, since the classifier acts as a linear hyperplane finding a separation of the representations in high dimensional space, increase in classifier variance may be limited by the representational feature imbalance in feature-shift non-IID settings. We introduce an additional regularization term to spread the representations of the encoder throughout the hypershpere. Fig. \ref{fig_feduv_method} is a visualization of the FedUV regularization scheme in $R^{2}$ (a PyTorch-like pseudocode is included in Appendix 1).
We first promote hyperspherical uniformity in the representations of the encoder (row vectors of the penultimate layer output) then promote the class-wise variance of the probability distribution (column vectors of the last layer output). We show that both regularization terms are important to increasing the performance in label-shift and feature-shift settings.

Our overall loss function is shown in Eq. \ref{local_obj_eq}. Each client uses the cross-entropy loss as well as two regularization terms. The $\mu$ and $\lambda$ terms control the strength of each regularization. These parameters will be discussed in Section \ref{experimental_setup}. We now discuss each regularization term in more detail.

\begin{equation}
    \label{local_obj_eq}
    \mathcal{L} =
     \mathcal{L}_{CE}(f_{\theta}(X), Y) 
     + \mu \mathcal{L}_{U}(g_{\theta}(X)) 
     + \lambda \mathcal{L}_{V}(f_{\theta}(X))
\end{equation}

\subsection{Classifier Variance}
Our goal is to encourage the classifier in any setting to emulate a classifier in an IID setting. We specifically focus on the classifier as it is the layer of the model most prone to bias in non-IID settings. 
We do this by inducing the class-wise variance of the classifier probability distribution to emulate an IID scenario. 

Given a mini-batch $(X,Y)$ and a model $f_{\theta}$ that outputs a prediction $f_{\theta}(X)=\hat{Y}$, we create a probability distribution on our mini-batch $\hat{P} := Softmax(\hat{Y})$. $\hat{P}$ is the probability distribution matrix of the entire mini-batch, where the row vectors are probabilities of a single sample and the column vectors are probabilities of a single class across the entire mini-batch. 
We define the variance regularization term $\mathcal{L}_V$ as the hinge loss of the square root of the variance (standard deviation) along the column vectors of the probability matrix as seen in Eq. \ref{var_loss_eq}.

\begin{equation}
    \begin{multlined}
    \label{var_loss_eq}
    \mathcal{L}_V(\hat{Y}) = 
    \frac{1}{D} \sum_{j=1}^{D} max(0, c - \sqrt{Var(\hat{P}_{j})})
    \end{multlined}
\end{equation}

\noindent Here, $D$ is the number of classes, $c$ is a constant that represents the variance in an IID setting, and $\hat{P}_{j}$ represents the column vectors of the probability matrix.
We set $c$ to be the average class-wise standard deviation of $\mathbf{A}$ as shown in Eq.~\ref{var_loss_forC_eq}. Here, $\mathbf{A}$ is a $D$x$D$ identity matrix ($diag(1_{1}, ..., 1_{D})$) where $D$ is the number of classes. This represents the ideal probability distribution of a balanced mini-batch where all classes are included. 


\begin{equation}
    \begin{multlined}
    \label{var_loss_forC_eq}
    c := \frac{1}{D} \sum_{j}^{D}{\sqrt{Var(\mathbf{A}_{j})}}
    \end{multlined}
\end{equation}


\noindent In essence, we ask that the predicted probability distribution in any setting match the probability distribution of an ideal IID setting where each mini-batch contains a balanced ratio of the included classes. 
Furthermore, we purposefully use a hinge loss to prevent the variance in a few classes from overpowering the remaining classes.

\subsection{Hyperspherical Uniformity}
Promoting class-wise variance in the classifier may be difficult in scenarios where the representations are clustered in a small area. For instance, in a feature-shift non-IID setting a model may biased towards a subset of the global data manifold.
This may also occur in scenarios where a client holds data from only two classes. There is no incentive for the model to spread the representations to allow larger separation with a linear hyperplane. We thus induce hyperspherical uniformity on the representations of the encoder (the output of the penultimate layer). 
We use the RBF kernel for its connection with the unit hypersphere \cite{rbfKernel_cohn2007universally}, and its success in various deep learning applications \cite{hUnif1_liu2018learning, unif_discuss0}. Namely, we calculate the average pairwise Gaussian energy between the row vectors of the representations as seen in Eq. \ref{unif_loss_eq}.

\begin{equation}
    \begin{multlined}
    \label{unif_loss_eq}
    \mathcal{L}_U(g_{\theta}(X)) =
    \mathop{\mathbb{E}}_{x,y \in g_{\theta}(X)}\left[\exp{(-\frac{\lVert x - y \rVert^{2}_{2}}{2\sigma} )}\right]
    \end{multlined}
\end{equation}

\noindent $g_{\theta}(X)$ is the representation matrix of the encoder where $x$ and $y$ represent the row vectors in the representations. $\sigma$ is a hyperparameter that controls the sensitivity to small distances. This is a hyperparameter that can be tuned to each setting to improve performance. We set $\sigma$ as the median value of the squared pairwise distances to ensure hypershperical uniformity is induced in any setting.

\section{Experiments and Results}
\label{sec:results}

\begin{table*}[t]
\centering
\setlength{\tabcolsep}{6pt} 
\renewcommand{\arraystretch}{1} 
\fontsize{8pt}{8pt}\selectfont

\begin{tabular}{l|cc|cc|cc|c|c|c}
\hline
& \multicolumn{2}{c|}{\multirow{2}{*}{STL-10}}
&\multicolumn{2}{c|}{\multirow{2}{*}{CIFAR-100}} 
&\multicolumn{2}{c|}{\multirow{2}{*}{Tiny ImageNet}}
&\multicolumn{1}{c|}{\multirow{2}{*}{PACS}}
&\multicolumn{1}{c|}{\multirow{2}{*}{HAM10000}}
&\multicolumn{1}{c}{\multirow{2}{*}{Office-Home}}\\
& \multicolumn{2}{c|}{}  & \multicolumn{2}{c|}{} & \multicolumn{2}{c|}{}
& \multicolumn{1}{c|}{}   & \multicolumn{1}{c|}{}  & \multicolumn{1}{c}{}\\
Method   & $\alpha$=0.01  & $\alpha$=1.0 &  
           $\alpha$=0.01    & $\alpha$=1.0 &  
           $\alpha$=0.01  & $\alpha$=1.0 &
          \hspace{15 mm} & \hspace{5 mm} & \hspace{5 mm} \\ \hline
         
FedAvg  & \underline{27.6$\pm$1.6} & 68.5$\pm$0.7 & \underline{51.6$\pm$1.7} & 58.3$\pm$1.2 & 37.5$\pm$0.9 & 40.2$\pm$1.2 
        & 61.9$\pm$0.6 & \underline{73.7$\pm$0.3} & \underline{42.2$\pm$1.1} \\

FedProx & 26.5$\pm$1.5 & 67.9$\pm$0.8 & 50.2$\pm$1.1 & 59.0$\pm$1.0 & 37.1$\pm$1.1 & 39.9$\pm$1.0 
        & 59.1$\pm$0.8 & 73.5$\pm$0.3 & 41.9$\pm$0.6 \\

MOON    & 26.0$\pm$1.3 & \underline{70.0$\pm$0.6} & 48.2$\pm$1.2 & \textbf{60.5$\pm$0.7} & 37.6$\pm$1.8 & 42.4$\pm$0.9 
        & \underline{62.8$\pm$1.6} & 72.9$\pm$0.4 & 41.9$\pm$0.4 \\

Freeze  & 23.7$\pm$1.5 & \textbf{72.1$\pm$0.6} & 51.1$\pm$1.5 & \underline{59.3$\pm$1.8} & \underline{38.9$\pm$1.9} & \underline{42.6$\pm$1.6}
        & 61.6$\pm$0.5 & 73.2$\pm$0.3 & 41.3$\pm$1.8 \\

FedUV   & \textbf{30.4$\pm$1.4} & 68.5$\pm$0.6 & \textbf{55.7$\pm$1.0}    & 59.1$\pm$0.9    & \textbf{40.3$\pm$0.8} & \textbf{43.2$\pm$1.5} & \textbf{65.9$\pm$0.9} & \textbf{73.9$\pm$0.5} & \textbf{45.4$\pm$1.0} \\ \hline
\end{tabular}
\caption{Test accuracies with $\alpha \in \{0.01,1.0\}$ on STL-10, CIFAR-100, and Tiny ImageNet and PACS, HAM10000, and Office-Home. We \textbf{bold} the highest performing method and \underline{underline} the second highest performing method. }
\label{tab:alphas_datasets}
\end{table*}

\subsection{Experimental Setup}
\label{experimental_setup}
We compare FedUV with the state-of-the-art FL algorithms: FedAvg \cite{fedavg_mcmahan2017}, FedProx \cite{fedprox_li2020federated}, and MOON \cite{moon_li2021model}. We also include a method \textit{Freeze} which freezes the final layer. This is similar to FedBABU \cite{fedbabu_oh2021fedbabu} without personalization. We use the PyTorch \cite{pytorch} library and follow the official implementations for all available work. 

\noindent \textbf{Label-shift Datasets.} 
\textbf{STL-10} \cite{dataset_stl10} is a dataset with 10 classes with balanced 5,000 training and 8,000 validation samples.
\textbf{CIFAR-100} \cite{dataset_cifar10_100} is a dataset with 100 classes with 60,000 training and 10,000 validation samples. 
\textbf{Tiny ImageNet} \cite{dataset_tiny_imagenet} is a dataset downsampled from the ImageNet \cite{imagenet} dataset with 200 classes with balanced 100,000 training and 10,000 validation samples. 
Basic data augmentation, random cropping and horizontal flipping, is used consistently throughout all datasets and methods. As with many previous work \cite{moon_li2021model, fedbabu_oh2021fedbabu, fedalign_mendieta2022local}, we use the Dirichlet distribution to simulate label-shift non-IID settings. The $\alpha$ term controls the extent of non-IIDness with $\alpha = 0$ being most non-IID and $\alpha = \infty$ being IID. We use $\alpha \in \{0.01, 1.0\}$. We split the training datasets into a 90-10 training-validation dataset and use the original validation set as a test set. 

\noindent \textbf{Feature-shift Dataests.} A more recent work \cite{li2021fedbn} has shown it is important also to focus specifically on feature-shift environments. To simulate feature-shift, we use 3 datasets which includes data from various domains. 
\textbf{PACS} \cite{li2017deeper} is a dataset with 7 classes in 9,991 images from four domains: Photo (1,670 images), Art painting (2,048 images), Cartoon (2,344 images), and Sketch (3,929 images). We randomly sampled an equal 280 images per domain (40 images per class in each domain) for our testing dataset (1,120 images in total).
\textbf{HAM10000} \cite{tschandl2018ham10000} is a dataset with 7 classes in 10,015 images from four domains. The domains are represented by different institutions which collected dermoscopic (skin disease) images. Each domain has 3363, 2259, 439, 3954 images respectively. Due to the imbalance in data, we took roughly 20\% from each domain for 1955 images for our testing set.  
\textbf{Office-Home} \cite{venkateswara2017deep} is a dataset with 65 classes in 15,588 images from 4 domains: Art (2,427 images), Clipart (4,365 images), Product (4,439 images), and Photo (4,357 images). We randomly sampled an equal 455 images per domain (7 images per class in each domain) for our testing dataset (1,820 images in total). We resize images from these datasets into 96x96x3 images and apply basic data augmentation (random cropping and horizontal flipping). 

A detailed overview of the distributions of the datasets are provided in Appendix 2.

\noindent \textbf{Models.} We use 3 different models across the different datasets. Note that we also add a non-linear projector following \cite{moon_li2021model} due to reduced performance without it. This is simply two fully-connected layers followed by a batch norm and ReLu activation. For STL-10 and PACS, we used a small CNN model (Appendix 3) with a projector with 256 neurons. For CIFAR-100 and PACS, we used a ResNet-18 \cite{resnet_he2016deep} model, removing the initial max-pooling layer due to the small image size and adding the projector with 512 neurons. For Tiny ImageNet and Office-Home, we used a ResNet-50 model, removing the initial max-pooling layer and adding the projector with 2048 neurons.
On all datasets, we report the results of a single aggregated global model. 

\noindent \textbf{Hyperparameters.} Our default testing environment for our ablation is done on STL-10 ($\alpha = 0.01$) with a small CNN (Appendix 3) and Office-Home with a ResNet-50, local epoch $E = 10$, number of clients $\kappa = 10$ for STL-10 and $\kappa = 4$ for Office-Home (for each of the domains), participation rate $\rho = 1.0$. The total aggregation rounds is chosen based on performance of the training-validation split on FedAvg. We set total aggregation rounds, $R$, as 100, 60, 40 for STL-10, CIFAR-100, Tiny ImageNet, and 60, 60, 40 for PACS, HAM10000, and Office-Home, respectively. All reported accuracy are averaged over 3 runs. Changes from these default settings are mentioned clearly. We use the Cross-entropy loss and SGD with learning rate of 0.01, momentum 0.9, and weight decay of 1e-5.

FedProx, MOON, and FedUV are all regularization methods, thus adds an extra parameter to balance the cross-entropy loss and the strength of regularization. We tune this parameter on the CIFAR-10 \cite{dataset_cifar10_100} dataset, since tuning this parameter in every setting of each dataset is not viable option for FL as resources are limited. 
For FedProx, we use $\mu = 0.01$, for MOON, $\mu = 1.0$. For FedUV, we use fixed hyperparameters $\lambda = \frac{D}{4}$, where $D$ is the number of classes, and $\mu = 0.5$ for all datasets. We report the hyperparameter tuning results in the appendix (Appendix 4). All experiments were conducted on a single RTX A5000 and one AMD EPYC 7763 processor.

\subsection{Results}
\label{actual_results}
\noindent \textbf{Data Heterogeneity.} 
We study how data heterogeneity changes performance of the global model. We test the performance for $\alpha \in \{0.01, 1\}$ for label-shift datasets, and use the defined domains for feature-shift datasets. Results are shown in Table \ref{tab:alphas_datasets}.

Across most settings, FedUV outperforms other methods. 
Focusing first on the label-shift non-IID settings, we see that in the extreme $\alpha = 0.01$ setting, FedUV performs best. FedProx and MOON peform worse than FedAvg as they base their regularization on the global model which may also be biased since it is averaging extremely biased local models. Freeze also does not perform well, as the frozen classifier causes the encoder to have difficulty learning in extreme settings.
By emulating an IID setting, FedUV outperforms these methods.

In the less extreme setting of $\alpha=1.0$ other regularization methods generally perform well. On STL-10, Freeze performs best by a large margin. For less extreme label shift and small number of classes, a frozen classifier is beneficial as this prevents the classifier from being biased, which has been established as a desirable property. 
MOON performs best on CIFAR-100, and also improves performance compared to FedAvg on the Tiny ImageNet dataset. However, we find that Freeze and FedUV achieve the better performance on Tiny ImageNet most likely because each client holds more data as the dataset is much larger. When data is plentiful, it is likely that focusing specifically on the classifier is most beneficial. 

In the feature-shift setting, we find that baseline regularization methods generally does not help performance. This could be because these work have focused only on label-shift settings. Freeze simply freezes the final layer, while MOON and FedProx use the global model as a source of regularization to mitigate the bias of individual clients. However, because FedUV simulates an IID setting rather than mitigating the effects of label-shift by relying on the global model as a source of regularization, performance is improved not only in label-shift settings but also feature-shift settings. As will be shown in the ablation study, the Uniformity regularizer is indeed an important factor to prevent this performance degradation. 

\begin{table}[t]
\centering
\fontsize{7pt}{7pt}\selectfont
\setlength{\tabcolsep}{3pt} 
\renewcommand{\arraystretch}{1.0} 
\begin{tabular}{l|ccc|ccc}
\toprule
        & \multicolumn{3}{c|}{STL-10 ($\alpha=0.01$)} & \multicolumn{3}{c}{Office-Home} \\
Method  & p=0.1           & p=0.5            & p=1.0              & p=0.25          & p=0.5  & p=1.0  \\ \midrule
FedAvg  & 17.4$\pm$3.4    & 21.5$\pm$1.8     & 27.6$\pm$1.6       & 39.1$\pm$2.5    &  41.9$\pm$1.4      & 42.2$\pm$1.1 \\
FedProx & 19.3$\pm$2.9    & 22.0$\pm$1.3     & 26.5$\pm$1.5       & 39.6$\pm$2.1    &  40.8$\pm$1.7      & 41.9$\pm$0.6 \\
MOON    & 20.5$\pm$3.7    & 23.6$\pm$1.7     & 26.0$\pm$1.3       & 40.3$\pm$2.7    &  40.5$\pm$1.6      & 41.9$\pm$0.4 \\
Freeze  & 21.5$\pm$3.6    & 23.7$\pm$1.6     & 23.7$\pm$1.5       & 37.6$\pm$3.9    &  39.5$\pm$1.1      & 41.3$\pm$1.8 \\
FedUV   & \textbf{24.9$\pm$3.1}    & \textbf{30.0$\pm$1.1}     & \textbf{30.4$\pm$1.4}       & \textbf{41.7$\pm$2.4}    &  \textbf{44.8$\pm$1.3}      & \textbf{45.4$\pm$1.0} \\ \bottomrule
\end{tabular}
\caption{Accuracy across client participation rate}
\label{tab:rho_participation}
\end{table}

\begin{figure*}[t]
\centering
\begin{subfigure}{.23\textwidth}
  \centering
  \includegraphics[width=.95\linewidth]{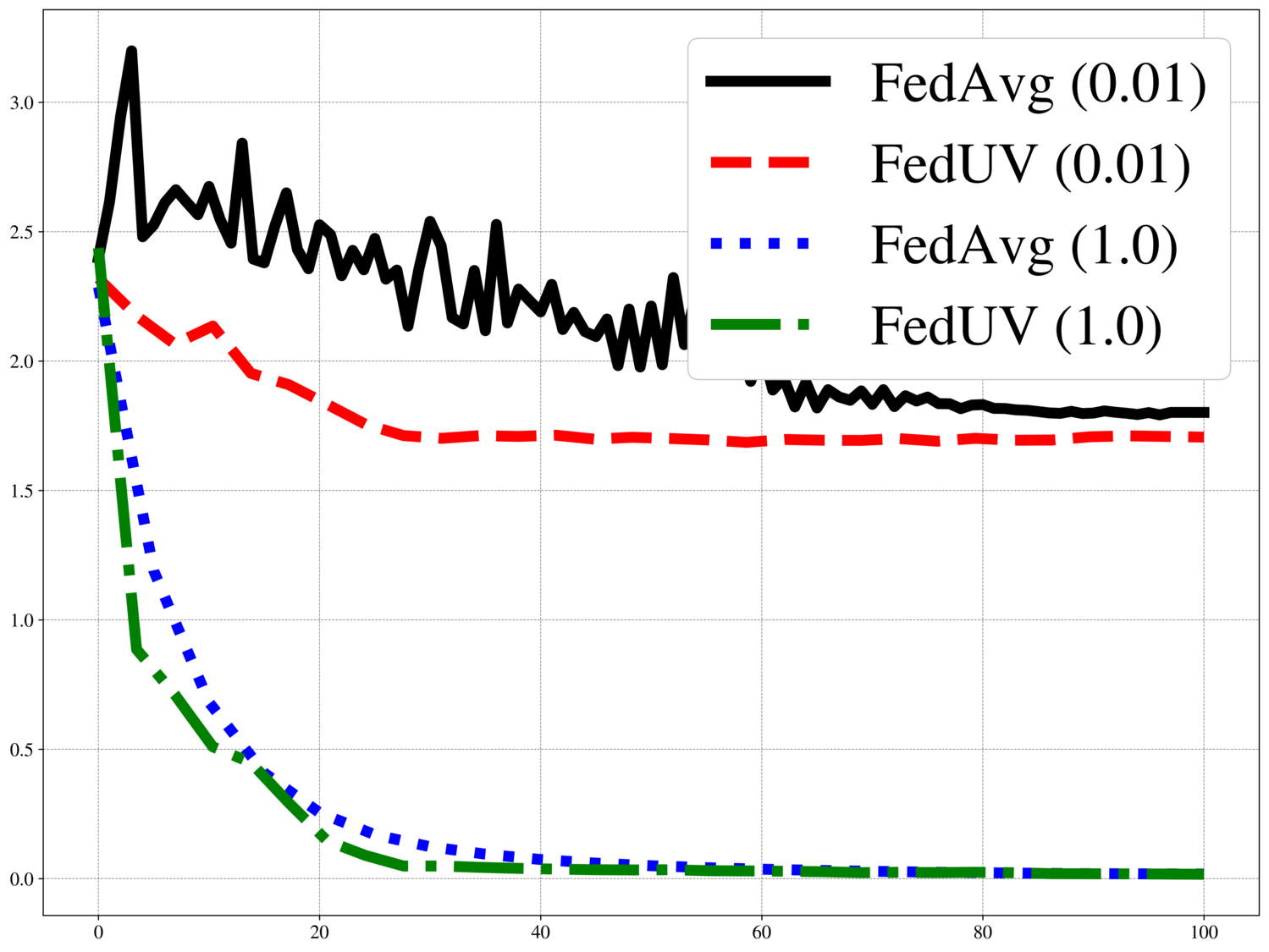}
  \caption{STL-10}
  \label{fig:sub1}
\end{subfigure}%
\begin{subfigure}{.23\textwidth}
  \centering
  \includegraphics[width=.95\linewidth]{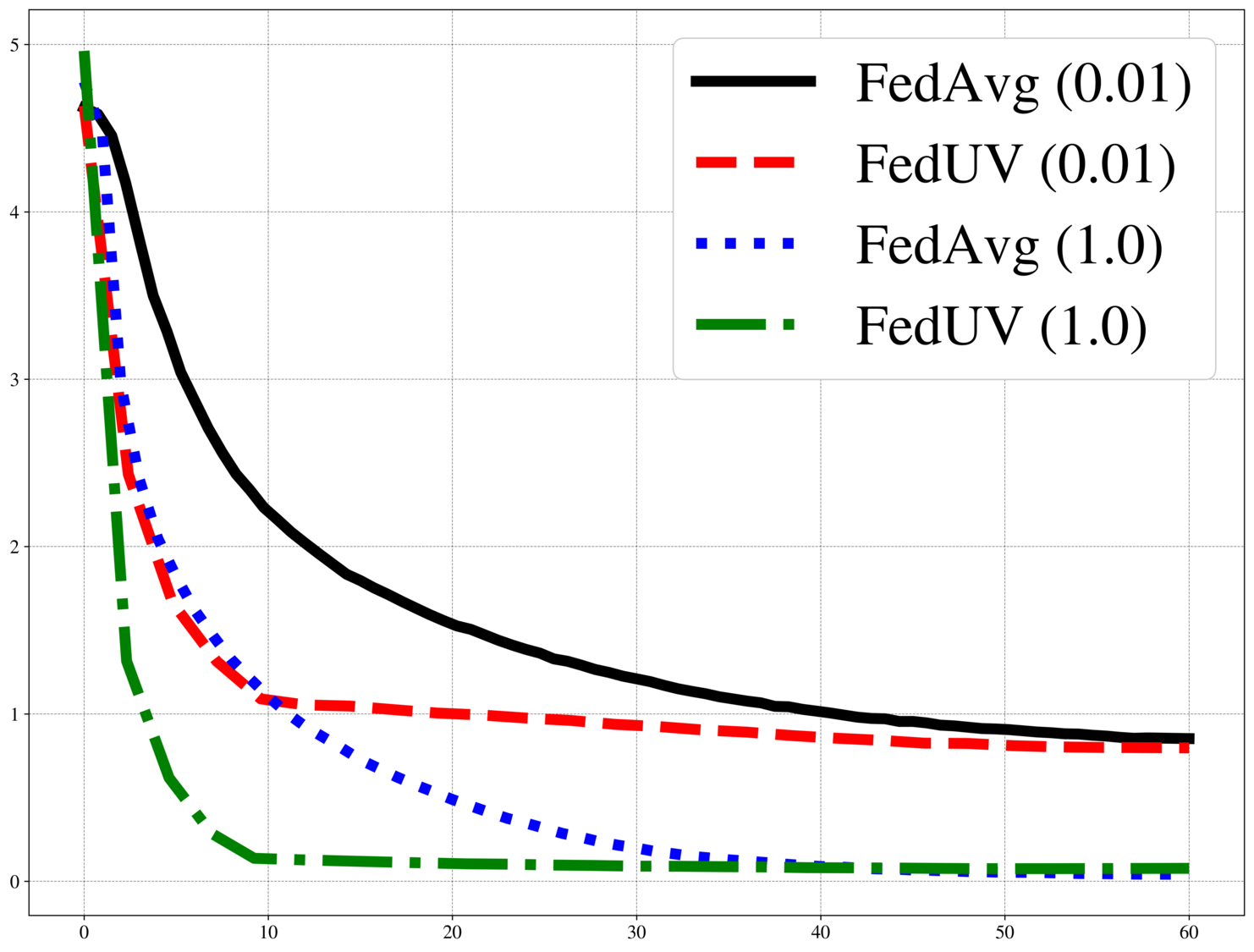}
  \caption{CIFAR-100}
  \label{fig:sub2}
\end{subfigure}
\begin{subfigure}{.23\textwidth}
  \centering
  \includegraphics[width=.95\linewidth]{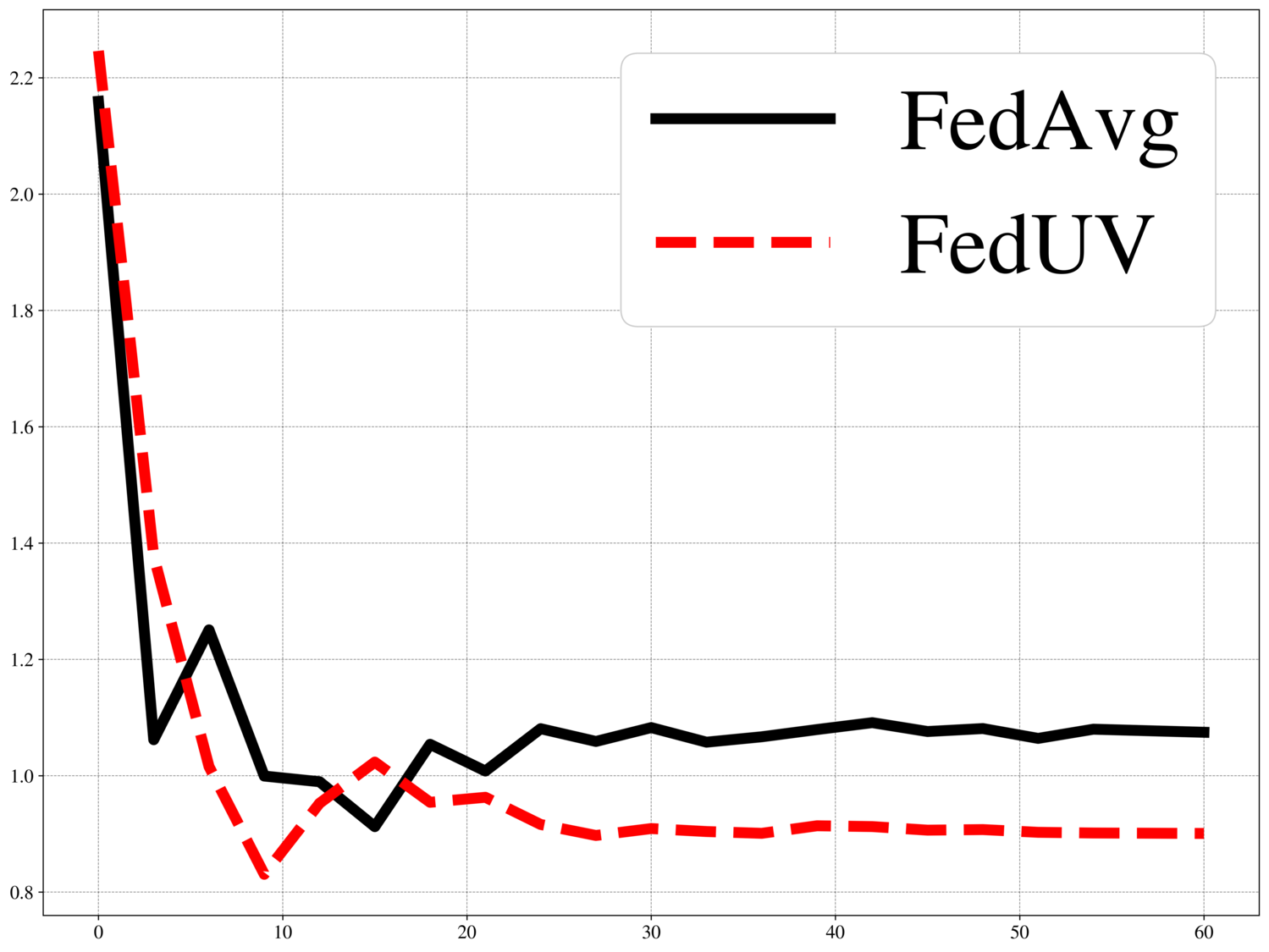}
  \caption{PACS}
  \label{fig:sub2}
\end{subfigure}
\begin{subfigure}{.23\textwidth}
  \centering
  \includegraphics[width=.95\linewidth]{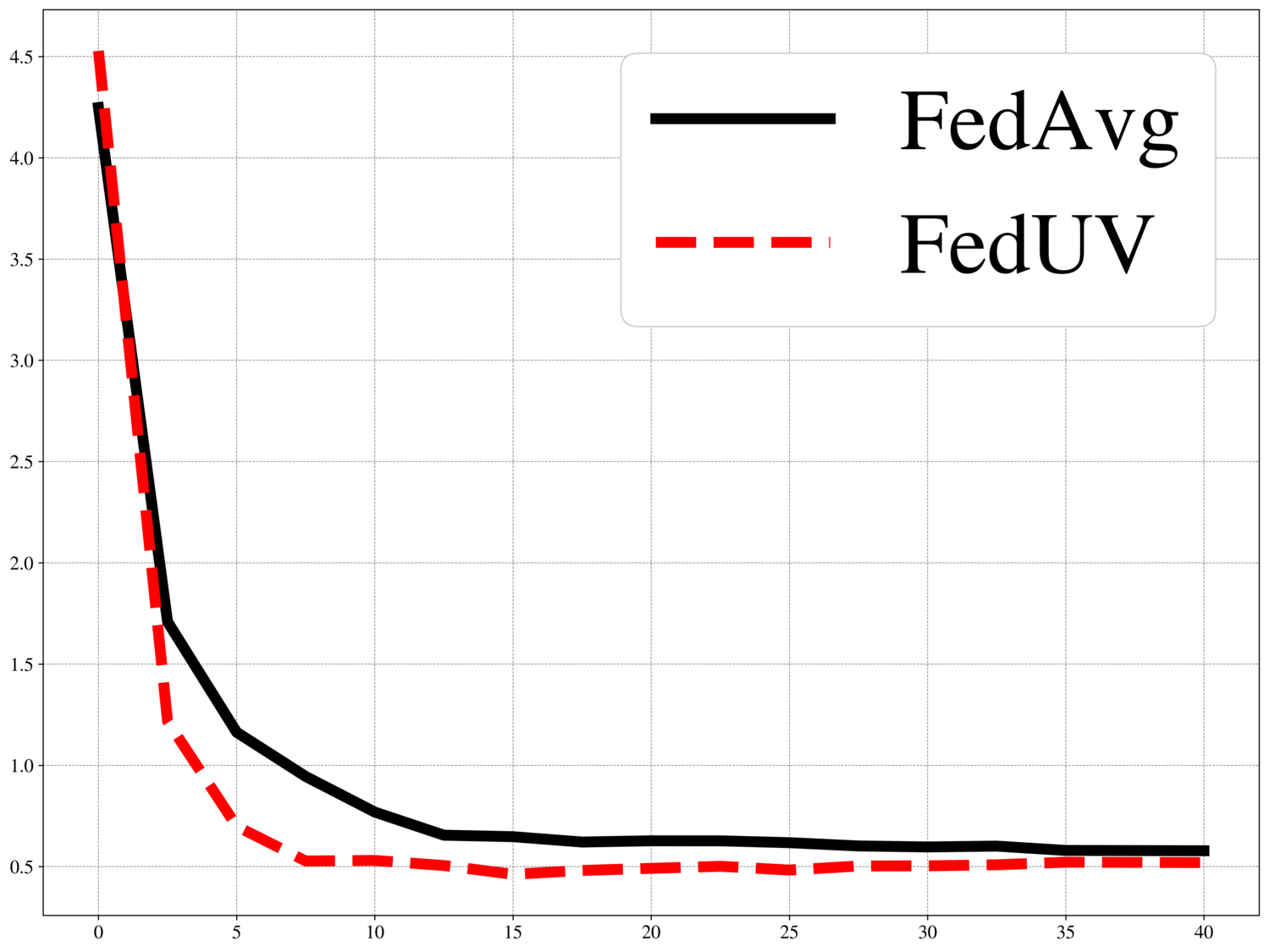}
  \caption{Office-Home}
  \label{fig:sub2}
\end{subfigure}
\caption{The training loss for FedAvg and FedUV across different settings}
\label{fig:fig3_trainingloss}
\end{figure*}

\noindent \textbf{Client participation.} 
In real-world FL applications, it is realistic to expect that only a subset of all clients participate in an aggregation round. The server must triggers the aggregation process even if some clients do not participate. 
We study the effect of ratio of client participation, $\rho$, where $\rho=0.1$ means 10\% of clients participate in the aggregation round. Across these settings, we use 100 aggregation rounds for STL-10 and 60 for Office-Home. The results are shown in Table \ref{tab:rho_participation}.

The advantages of FedUV are clear even when the ratio of participating clients is low. Indeed, when $\rho=0.1$ for STL-10, FedUV performs the best by a large margin. This increase in performance occurs owing to the focus of FedUV to emulate the IID. The second best method is Freeze. Intuitively, since the classifier is the layer that is most prone to bias, this is magnified in scenarios where client participation is low, as bias of the averaged global model is further increased. By fixing the classifier as in Freeze, we limit this bias and increase performance. However, as seen with the performance increase of FedUV, it is far more beneficial to emulate an IID setting. It is also interesting to see that when $\rho=0.25$ for Office-Home (1 client participating), FedProx and MOON improves performance. When the server aggregates weights from only 1 domain, it may be beneficial to use the global model as regularization. This improvement does not continue for $\rho=0.5$ (2 clients participating), suggesting that the average of two domains is more beneficial than the use of the global model for regularization.

\begin{table}[t]
\centering
\fontsize{7pt}{7pt}\selectfont
\setlength{\tabcolsep}{3pt} 
\renewcommand{\arraystretch}{1.0} 
\begin{tabular}{l|ccc|ccc}
\toprule
        & \multicolumn{3}{c|}{STL-10 ($\alpha=0.01$)} & \multicolumn{3}{c}{Office-Home} \\
Method  & $E=10$           & $E=20$            & $E=40$             & $E=10$          & $E=20$  & $E=40$  \\ \midrule
FedAvg  & 27.6$\pm$1.6    & 29.8$\pm$1.1     & \textbf{29.1$\pm$0.3}       & 42.2$\pm$1.1    & 43.5$\pm$1.0     & 35.1$\pm$0.7   \\
FedProx & 26.5$\pm$1.5    & 28.6$\pm$0.8     & 27.4$\pm$0.6       & 41.9$\pm$0.6    & 43.3$\pm$0.5     & 35.7$\pm$0.5       \\
MOON    & 26.0$\pm$1.3    & 28.9$\pm$0.7     & 27.4$\pm$0.4       & 41.9$\pm$0.4    & 42.0$\pm$0.4     & \textbf{41.7$\pm$0.5}       \\
Freeze  & 23.7$\pm$1.5    & 29.9$\pm$0.9     & 27.2$\pm$0.7       & 41.3$\pm$1.8    & 41.3$\pm$0.8     & 35.4$\pm$0.9       \\
FedUV   & \textbf{30.4$\pm$1.4}    & \textbf{31.5$\pm$1.2}     & 25.9$\pm$0.9       & \textbf{45.4$\pm$1.0}    & \textbf{46.1$\pm$0.7}     & 35.9$\pm$0.8       \\ \bottomrule
\end{tabular}
\caption{Accuracy across different number of local epochs}
\label{tab:E_localEpochs}
\end{table}

\noindent \textbf{Number of local epochs} 
We study the effect of number of local epochs, $E$, where $E = 40$ means each clients trains for 40 epochs during a single aggregation round. Results are shown in Table \ref{tab:E_localEpochs}. It would be desirable to achieve high performance with a low number of local epochs thereby reducing the burden of less powerful edge devices. We find that when $E = 20$ performance generally improves.
However, when the number of local epochs increase to 40, the accuracy of all methods decrease, suggesting overfitting occurs with a high number of local epochs. 
The performance of FedUV significantly degraded when local epoch is high. This suggests high local epochs cause local models to converge to local optima, harming performance. On the other hand, MOON and FedProx did not show a large drop in performance. This can be explained by the fact that these methods rely on the global model for regularization. Because of this, local models are less likely to converge to local optima as compared to FedUV.  

\begin{table}[t]
\centering
\fontsize{7pt}{7pt}\selectfont
\setlength{\tabcolsep}{3pt} 
\renewcommand{\arraystretch}{1.0} 
\begin{tabular}{l|ccc|ccc}
\toprule
        & STL & CIFAR & Tiny & PACS & HAM & Office \\
Method  &      &              &             &         &  &   \\ \midrule
FedAvg        & 27.6$\pm$1.6  & 51.6$\pm$1.7 & 37.5$\pm$0.9 & 61.9$\pm$0.6 &  73.3$\pm$0.3  & 42.2$\pm$1.1      \\
Freeze        & 23.7$\pm$1.5  & 54.1$\pm$1.5 & 38.9$\pm$1.9 & 61.6$\pm$0.5 &  73.2$\pm$0.3  & 41.3$\pm$1.8      \\
Only $L_{V}$  & 29.5$\pm$1.4  & 54.0$\pm$1.1 & 38.8$\pm$1.5 & 60.1$\pm$0.3 &  72.5$\pm$0.4  & 41.8$\pm$1.1       \\
Only $L_{U}$  & 24.1$\pm$1.3  & 52.8$\pm$1.2 & 37.9$\pm$1.7 & 63.7$\pm$0.3 &  73.4$\pm$0.2  & 43.5$\pm$1.3     \\
FedUV         & 30.4$\pm$1.4  & 55.7$\pm$1.0 & 40.3$\pm$0.9 & 65.9$\pm$0.9 &  73.9$\pm$0.5  & 45.4$\pm$1.0       \\ \bottomrule
\end{tabular}
\caption{FedUV ablation study}
\label{tab:ablations}
\end{table}

\noindent \textbf{Ablation Study.} 
We study the effect of hyperspherical uniformity and classifier variance separately. We also compare Freeze which creates an orthogonal classifier by randomly initializing and freezing the weights. Results are shown in Table \ref{tab:ablations}.
For the label-shift datasets (STL-10, CIFAR-100, Tiny ImageNet), we find that hyperspherical uniformity regularization, \textit{Only $L_{U}$}, does not improve performance significantly, whereas classifier variance regularization, \textit{Only $L_{V}$}, performs much better than Freeze on the STL-10 dataset while being competitive in other datasets. These results suggests that label-shift settings, focusing on the final layer is more important than focusing on the feature representations of the encoder, and that emulating an IID label distribution is more desirable than a fixed classifier. 

Furthermore, for the feature-shift datasets (PACS, HAM10000, Office-Home), we find that varaince regularization does not improve performance significantly, whereas uniformity regularization improves performance compared to both \textit{Only $L_{V}$} and Freeze. This suggests that in feature-shift settings, when clients have vastly differing feature distributions, it is beneficial to encourage features to be more spread. Intuitively, this prevents any local encoder from heavy bias towards a subspace, which is an aspect other regularization methods do not incorporate.

Overall, these results suggests that, in label-shift situations, focusing specifically on the classifier through variance regularization is more important, while in feature-shift scenarios, focusing specifically on the feature space representations of the encoder through hyperspherical uniformity is more important. However, both regularization terms are important to emulate the IID setting regardless of data distribution. 
The combined approach in FedUV shows further improved performance across these various settings.

\noindent \textbf{Convergence.} We study the convergence of FedUV. Graphs of the training loss are shown in Fig.~\ref{fig:fig3_trainingloss}. FedUV shows quicker convergence when compared to FedAvg, often reaching a lower loss at earlier training iterations. This is intuitive considering the performance improvement FedUV provies over FedAvg. We also find that in the extreme case of STL-10 $\alpha=0.01$, the loss curve is much smoother for FedUV when compared to FedAvg. This shows FedUV stabilizes training.
\section{Discussion}
\label{sec:discussion}

\begin{figure*}[hbt!]
\centering
    \includegraphics[width=0.75\textwidth,trim=1 1 1 1,clip]{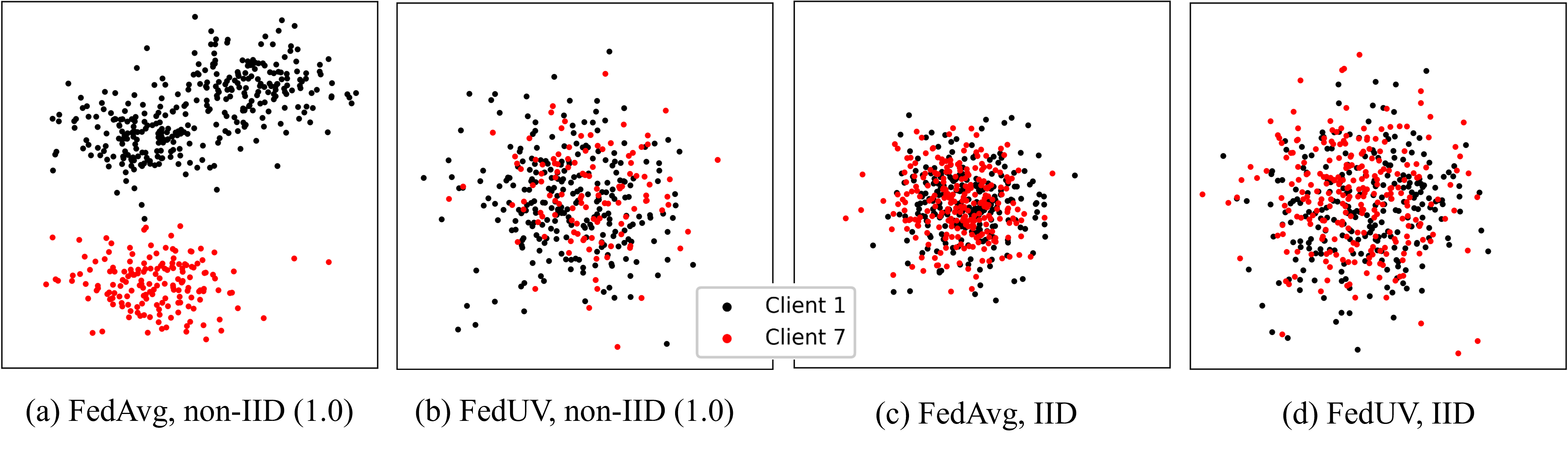} 
\caption{t-SNE on the features (output of the penultimate layer) of STL-10 (Class \#1). Experiment setup is equivalent to main text. Clients and class were selected at random.} 
\label{fig_tsne}
\end{figure*}

\subsection{Preventing Classifier Bias}
Our method of emulating the IID setting by encouraging classifier variance is not the only method to prevent classifier bias. When focusing on label-shift, many loss function weighting and specialized sampling techniques have been considered \cite{classImba_survey}. There have also been interesting applications in representation learning that use latent space representations to sample training data such that the predicted probability distribution is uniform \cite{amini2019uncovering}. These are interesting directions to explore for addressing the non-IID problem.
The biggest contraint using these approaches in FL however, is privacy. The data and class distributions cannot be known beforehand, thus cannot be set to match each client. In addition, parameters that control sampling or loss weighting may introduce security vulnerabilities.  

Another straightforward method would be to randomly initialize and freeze the classifier. 
Freezing the classifier would also prevent biases forming in the probability distribution of the classifier. Indeed, this is the intuition behind FedBABU \cite{fedbabu_oh2021fedbabu} and the Freeze method we have used as our baseline, which initializes and freezes the classifier to orthogonal vectors.
Freezing weights also prevents the degeneration of singular values in non-IID settings, a phenomenon shown in Fig.~\ref{fig_svdcomp}.

However, it may not be optimal to separate classes with orthogonal vectors in every setting. 
A number of work \cite{hUnif0_liu2021learning, hUnif1_liu2018learning} have shown that when the number of neurons exceeds the dimensions of data manifold, promoting orthogonality is problematic. This problem arises when there is a large number of classes with data that lives in a lower subspace.
Another alternative would be to regularize the weights directly. One could maximize the spectral norm, the largest singular value, or minimize the Frobenius norm, the ${L_{2}}$ vector-norm of the singular values. These methods may also bridge the difference in singular values between IID and non-IID while preventing classifier basis. These methods differ from FedUV as we regularize in the representation space (the outputs of the model) rather than in the weight space. The main reason is efficiency. We re-use the representations obtained in the forward-pass of the model without directly accessing the weight parameters. 


\subsection{Hyperspherical Uniformity}
The goals of hypershperical uniformity in FedUV are to mitigate representational bias and to create separation between representations such that the classifier has more directions to increase its variance. Note that this regularizer focuses on feature distributions, not class distributions. This is shown in Figure \ref{fig_tsne}. Features are aligned regardless of class distribution. Empirically, we have shown that this regularization is beneficial in feature-shift settings.
There are alternatives to this, such as spreading representations along a hypercube. However, hyperspherical uniformity is gaining popularity in the field of deep learning due to its desirable properties \cite{hUnif1_liu2018learning, unif_discuss0, unif_discuss1, unif_discuss2, unif_discuss3}. While more traditional regularization approaches such as the $L_{2}$ regularization have seen success in various applications in the wider field of machine learning and statistics \cite{perceptual0_johnson2016perceptual, perceptual1_gatys2016image}, neural networks are special due to their high over-parameterization. Using $L_{2}$ regularization essentially focuses only on individual weights rather than taking into account the interactions between these over-parameterized neurons. Hyperspherical uniformity regularizes based on these interactions and also has clear geometric properties. We point readers to \cite{hUnif0_liu2021learning} for a recent in-depth discussion on its development in deep learning.

Furthermore,  we choose the RBF kernel for its efficiency. In particular, the kernel method allows us to first calculate the the pairwise ${L_{2}}$ norms of the row vectors before using the kernel function. It also allows for easy differentiation on automatic differentiation software, which facilitate the training of neural networks.

\begin{table}[t]
\centering
\fontsize{7pt}{7pt}\selectfont
\setlength{\tabcolsep}{1.5pt} 
\renewcommand{\arraystretch}{1.0} 
\begin{tabular}{l|cccccc}
\toprule
        & STL & CIFAR & Tiny & PACS & HAM & Office \\
Method  & (10-Layer) & (ResNet-18) & (ResNet-50) & (10-Layer) & (ResNet-18) & (ResNet-50) \\ \midrule
FedAvg   & 26  & 57  & 728  & 47 & 131 & 359      \\
FedProx  & 27  & 67  & 823  & 48 & 135 & 371     \\
MOON     & 37  & 88  & 1186 & 70 & 145 & 493     \\
Freeze   & 26  & 56  & 728  & 47 & 129 & 351     \\
FedUV    & 27  & 58  & 755  & 48 & 132 & 362      \\ \bottomrule
\end{tabular}
\caption{Average time (in seconds) per aggregation round.}
\label{tab:efficiency}
\end{table}

\subsection{Convergence of FedUV}
We have shown in Fig.~\ref{fig:fig3_trainingloss} that in our extensive testing environments, FedUV converges stably.
In the general case, we know that the Uniformity regularizer converges weakly to the uniform distribution, since we are minimizing the average pairwise gaussian potential between sample points. As the number of samples approaches infinity, the loss will converge weakly to the uniform distribution according to \citet{wang2020understanding_alignunif}.
For the Variance regularizer, however, the hinge loss of the regularizer is not smooth. Because of this, it is non-trivial to prove convergence. This may become much simpler when replaced with a smooth surrogate loss. We leave the proof for the convergence of this regularizer for future work.

\subsection{Efficiency and Scalability}
One of the most important aspects in FL is efficiency. Edge devices cannot be expected to have powerful hardware that allows heavy computation in real-world scenarios. We unfortunately find that many regularization methods are not very scalable. While this is not a problem for small datasets and models, as dataset size and model size both increase at a rapid rate, these models become more computationally expensive.

We compare the wall-clock time of different methods as shown in Table \ref{tab:efficiency}. Note that the time extended is per aggregation round with 10 local epochs. FedProx requires $L_{2}$ norm calculations between each layer of local model and global model at each batch, while MOON requires 3 forward passes at each batch as well as the storage of 3 full models. Though this may not be a problem with small models, FedProx and MOON are not very scalable methods as the size of models increases.
FedUV is a simple yet efficient regularization technique as it only requires few matrix operations with no weight parameter access, extra memory, or extra forward passes. The most expensive operation is the pairwise distance calculation which is negligible even with less powerful hardware. 
\section{Conclusion}
\label{sec:conclusion}
In this work, we introduced a new approach to address the non-IID problem in FL by encouraging local models to emulate the IID setting. Rather than regularizing to reduce bias, we emulate the IID setting by promoting variance in the predicted dimension-wise probability distribution. We also promote hyperspherical uniformity on the representations of the encoder to allow the classifier to increase variance in more directions. Overall, our method improves performance by a large margin throughout our extensive experiments while being the most efficient and scalable among regularization methods.
{
    \small
    \bibliographystyle{ieeenat_fullname}
    \bibliography{main}
}


\end{document}